\pdfoutput=1

\documentclass[11pt]{article}

\usepackage[]{acl}

\usepackage{times}
\usepackage{latexsym}

\usepackage{graphicx}
\usepackage{pdfpages}
\usepackage{amsmath}
\usepackage{multirow}
\usepackage{hyperref}

\usepackage[T1]{fontenc}

\usepackage[utf8]{inputenc}

\usepackage{microtype}

\title{Overview of the SV-Ident 2022 Shared Task on \\
Survey Variable Identification in Social Science Publications}

\author{Tornike Tsereteli\textsuperscript{1}, %
Yavuz Selim Kartal\textsuperscript{2}, %
Simone Paolo Ponzetto\textsuperscript{1},\\%
\textbf{Andrea Zielinski\textsuperscript{3}}\textbf{,} \textbf{Kai Eckert\textsuperscript{4}}\textbf{,} \textbf{Philipp Mayr\textsuperscript{2}} \\%
  \textsuperscript{1}Data and Web Science Group, University of Mannheim, Germany \\
  \textsuperscript{2}GESIS -- Leibniz Institute for the Social Sciences, Germany \\%
  \textsuperscript{3}Fraunhofer ISI, Germany \\ 
  \textsuperscript{4}Web-based Information Systems and Services, Stuttgart Media University, Germany\\
  \texttt{\{tornike.tsereteli, ponzetto\}@uni-mannheim.de} \\ \texttt{\{yavuzselim.kartal, philipp.mayr\}@gesis.org} \\
  \texttt{andrea.zielinski@isi.fraunhofer.de} \\
  \texttt{eckert@hdm-stuttgart.de}}

\begin{document}
\maketitle

\begin{abstract}
In this paper, we provide an overview of the SV-Ident shared task as part of the 3rd Workshop on Scholarly Document Processing (SDP) at COLING 2022.
In the shared task, participants were provided with a sentence and a vocabulary of variables, and asked to identify which variables, if any, are mentioned in individual sentences from scholarly documents in full text.
Two teams made a total of 9 submissions to the shared task leaderboard.
While none of the teams improve on the baseline systems, we still draw insights from their submissions.
Furthermore, we provide a detailed evaluation.
Data and baselines for our shared task are freely available at \url{https://github.com/vadis-project/sv-ident}.
\end{abstract}

\section{Introduction}
\label{sec:introduction}
\begin{figure}
    \centering
    \includegraphics[width=\linewidth]{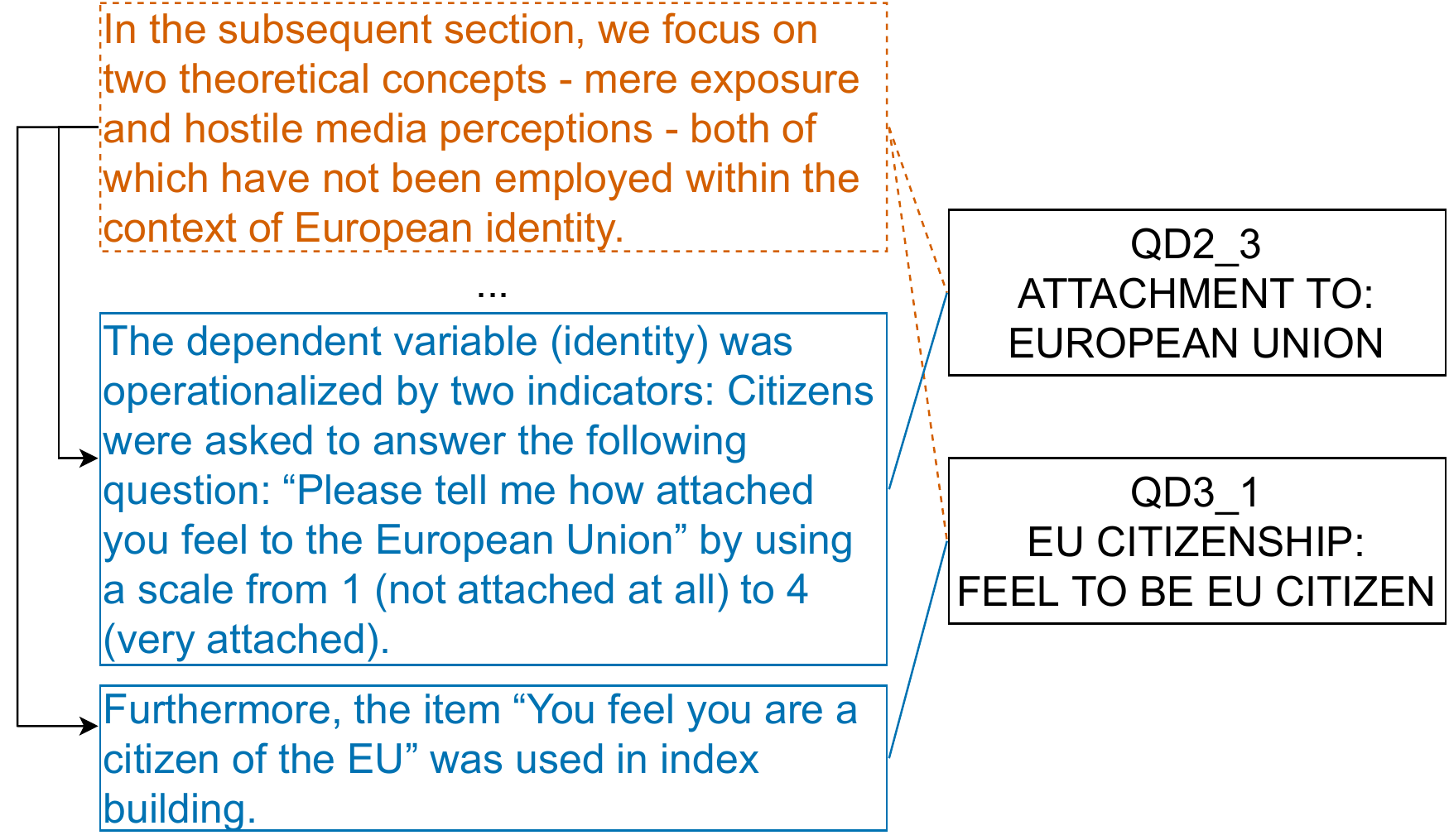}
    \caption{Example of {\color[HTML]{0072B2}explicit} ({\color[HTML]{0072B2}blue}) and {\color[HTML]{D35F00}implicit} ({\color[HTML]{D35F00}red}) variable mentions in sentences from a social science article (source:  \citet{MaC885}) mapped to survey variables. Lines with arrows show contextual dependence. Linked variables: \href{https://search.gesis.org/variables/exploredata-ZA5876_Varqd2_3}{QD2\_3} and \href{https://search.gesis.org/variables/exploredata-ZA5876_Varqd3_1}{QD3\_1}.}
    \label{fig:example}
\end{figure}

Social science publications often use and reference survey datasets, containing hundreds or thousands of questions, using so-called \textit{survey variables}.\footnote{In the following, we use the terms \textit{survey variable} and \textit{variable} interchangeably.}
While publications may focus only on a specific subset of these variables, explicit references are usually missing: the lack of explicit links between survey variables and publications, in turn, limits access to research along the FAIR principles \citep{Wilkinson2016}.
To address this issue, we propose a task where variable mentions in unstructured documents are linked to items from a catalog of survey research datasets using Natural Language Processing (NLP) methods.
Automatically identifying which variable is mentioned in a given text is challenging due to the diverse linguistic realizations of variables \cite{zielinski-mutschke-2018-towards}.
A short example text is shown in Figure~\ref{fig:example}.
All three sentences mention and are linked to relevant variables.
The first sentence mentions three concepts:\footnote{Concepts that have been operationalized by variables are also treated as variables throughout this work.} \textit{mere-exposure effect}, \textit{hostile media perceptions}, and \textit{European identity}.
The first two concepts are defined later in the text (we omit their links in this example), while the latter is defined in the bottom two sentences in the figure.
The second and third sentences both are explicit mentions, as they include direct quotations of variable questions.
Ideally, a system should link relevant variables to each of the sentences in the example.
Specifically, when only provided the given context, it should link the first sentence to the variables \texttt{QD2\_3} and \texttt{QD3\_1}, the second sentence to \texttt{QD2\_3}, and the third sentence to \texttt{QD3\_1}.
A larger variant of the example is provided in Figure~\ref{fig:large-example} in the Appendix.

The Survey Variable Identification\footnote{\url{https://vadis-project.github.io/sv-ident-sdp2022/}} (henceforth, SV-Ident) shared task aims at promoting the developing of systems that can identify variables within the text of scholarly publications from the social sciences in different languages (initially, we focus here on English and German).
The shared task is divided into two sub-tasks: a) Variable Detection and b) Variable Disambiguation.
The former deals with identifying sentences that contain variable mentions, while the latter focuses on linking the correct variables mentioned in a sentence.
Variable mentions are often implicit (e.g., sentences 1 and 3 in Figure~\ref{fig:example}), and understanding when a variable is mentioned may require contextual information as well as knowledge from external sources (e.g., a variable vocabulary).
Since annotating scientific texts requires domain knowledge, training data is costly to create and thus scarce.
To overcome these limitations, NLP systems, e.g., models using pre-trained language models (PLMs) and transfer learning are promising technologies to use.

In this paper, we report the results on the first edition of the SV-Ident shared task.
Two teams made a total of 9 submissions to the leaderboard.
One of the teams developed systems for both sub-tasks and submitted a system description paper.
While none of the teams improve on the baselines, we use the submissions provided by the teams to collect a few initial findings on the difficulties and challenges of the SV-Ident task.
Crucially, we find that there is a difference between the performance on two types of variable mentions: explicit and implicit.
Implicitly mentioned variables (sentence 1 in Figure~\ref{fig:example}) are significantly more difficult to detect and disambiguate, as they require contextual knowledge.
This opens up new research questions for future work, such as, for instance: can implicit mentions of survey variables be further categorized into finer-grained classes or can co-reference resolution be used to link variable mentions across different parts of a document?
In order to foster future research on this task, we release all of our code to reproduce the analysis results and the annotation guidelines for creating the dataset at \url{https://github.com/vadis-project/sv-ident}.

The remainder of this paper is organized as follows: we provide an overview of the dataset used in \S\ref{sec:data}.
In \S\ref{sec:experimental-setup}, we describe the task definition and evaluation metrics.
We present the submitted systems in \S\ref{sec:participating-systems} and provide a detailed analysis of the results in \S\ref{sec:evaluation}.
We briefly discuss related work in \S\ref{sec:related-work} and frame the shared task into a broader context in \S\ref{sec:why-sv-ident}.
Finally, we summarize the shared task and propose future work in \S\ref{sec:conclusion}.

\section{Data}
\label{sec:data}
The SV-Ident 2022 shared task has been conceived in the context of the VADIS project\footnote{\url{https://vadis-project.github.io/}} and organized as part of the third Workshop on Scholarly Document Processing (SDP) \cite{chandrasekaran-etal-2020-overview}, co-located at the 2022 International Conference on Computational Linguistics.
In the following, we describe the data collection process and the dataset used for the shared task.

\subsection{SV-Ident Corpus}
The SV-Ident Corpus contains publicly-available scientific publications from the Social Science Open Access Repository (SSOAR)\footnote{\url{https://www.gesis.org/ssoar/home}} in full text.
To collect the corpus, we first filter the 5,000 most popular research datasets using search logs from GESIS Search.\footnote{\url{https://search.gesis.org/}}
We then retrieve the publications linked to these datasets as our candidate set.
Finally, only those publications that had at least one associated research dataset with indexed variables on the GESIS Search platform are retained.
This results in 285 documents from the original set of 120k publications.
For this set of candidates, we then select 44 documents for annotation, which include the most popular ones as well as publications linked to variable vocabularies of different sizes.

Each document in our dataset has been annotated in PDF format using the INCEpTION software \citep{klie-etal-2018-inception} by two domain experts (graduate students trained in the social sciences).
Annotators are provided with the whole document and asked to label all sentences that contained variables, including the variables the sentences mentioned.
We first conduct two calibration rounds, for which annotators are given 50 two-page documents from the dataset collected by \citet{zielinski-mutschke-2018-towards}.
Afterwards, the selected 44 documents are annotated in three annotation rounds over a period of 8 weeks (on average, each annotator spent between 1-2 hours on each document).
Texts are then extracted, and all parsing errors are manually corrected.
Common errors include sentence breaks (due to incorrect splitting of abbreviations, such as \textit{et al.} or \textit{i.e.}), page breaks (due to improper handling of footnotes), and missing spaces between words.
Because annotators have access to all parts of the document at once, the annotation setup allows the use of document-level knowledge to infer sentence-level labels.

The annotations include the variable IDs that are mentioned in a text from a set of possible candidates, confidence scores for the annotations, and, for the test set, annotators also classified each mentioned variable into an \textit{explicit} or an \textit{implicit} mention (examples of explicit and implicit mentions were both found in the annotation guidelines).
We generally define explicit mentions as those which do not require contextual information to be labeled correctly.
The opposite is true for implicit mentions.

\subsection{SV-Ident Shared Task Dataset}
\begin{figure}
    \centering
    \includegraphics[width=\linewidth]{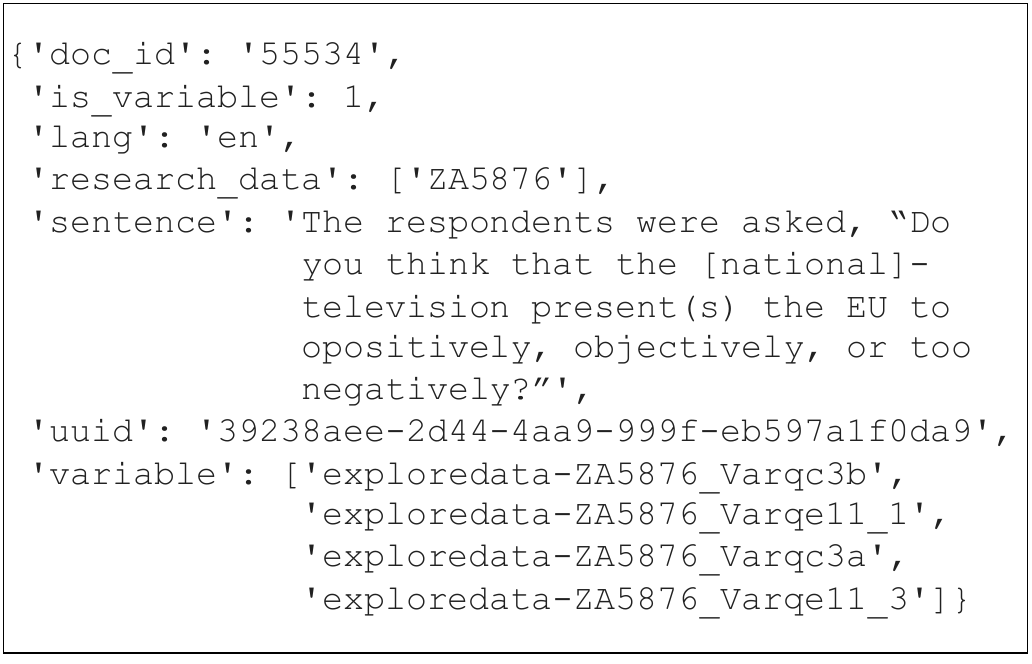}
    \caption{Example sentence with provided metetadata and labels.}
    \label{fig:instance}
\end{figure}

When annotating, the set of candidate variables is potentially made up of all variables from the research datasets linked to a publication on the GESIS platform: this set usually contains hundreds or thousands of variables, thus making the annotation task impractical and hard to scale.
To help reduce the size of the set of possible survey variable labels, annotators are provided with a tool to find matches using different methods.
The first method uses an ensemble of four sentence-transformer models to predict the top 20 variables that are semantically most similar to the reference sentence for each model.
The annotators receive recommendations for variables for which at least two models predict them to be in the top 20 results.
The second method allows annotators to search using a method of matching strings approximately rather than exactly: specifically, we use the \textit{Token Set Ratio} metric, which compares the number of insertion and deletion operations for unique and common words between the strings to be compared.\footnote{We use the RapidFuzz library (\url{https://github.com/maxbachmann/RapidFuzz}) to match relevant variables given a search query.}
The last method simply provides annotators with the full list of variables to manually search through.
All three methods have their drawbacks.
The first two might fail to recommend valid variables for cases with high linguistic variation, vagueness, or infrequent words, while the last may provide annotators with a search space that is too large.
While we do not control for such possible failures, future work may draw insights from the analysis of the annotations.

The dataset for the shared task is a subset of the SV-Ident corpus.
More specifically, 14 out of the 44 annotated documents from the SV-Ident Corpus are additionally filtered out due to missing links to research data, incorrect annotations, or PDF parsing errors, leaving 30 documents in total.
The dataset consists of 18 documents (7 English and 11 German) for the training and development sets and 12 documents (6 for each English and German) for the blind test set.

An example of a sentence and its metadata, including annotated labels from the dataset, is shown in Figure~\ref{fig:instance}.
Each instance in our dataset contains: a document ID (\textit{doc\_id}); a binary label (\textit{is\_variable}), where a value of \textit{1} implies that the sentence contains a variable; the language of the sentence (\textit{lang}); a list of document-level linked research datasets (\textit{research\_data}); the sentence (\textit{sentence}); a unique ID (\textit{uuid}); and a list of annotated variables (\textit{variable}).
Raw sentence counts for each of the dataset splits are provided in Table~\ref{tab:data_stats}.
Since the test set contains more English sentences, during evaluation, we compute the mean of the scores for each language for the competing systems (see \S\ref{sec:evaluation} for more details).
In total, there are 3,823 training, 425 validation, and 1,724 test sentences.
English and German sentences are roughly evenly distributed at 3,035 and 2,937 instances for each language, respectively.

\begin{table}[]
    \centering
    \begin{tabular}{l|r|r|r}
         & English & German & Total \\
        \hline
        \hline
        Train & 1,882 & 1,941 & 3,823\\
        Dev & 209 & 216 & 425\\
        Test & 944 & 780 & 1,724\\
        \hline
        Total & 3,035 & 2,937 & 5,972 \\
        \hline
    \end{tabular}
    \caption{Total number of sentences in the SV-Ident shared task dataset per language for each dataset split.}
    \label{tab:data_stats}
\end{table}

\begin{table}[]
    \centering
    \begin{tabular}{p{3cm}|r|r}
        Task / Metric & English & German \\
        \hline
        \hline
        Detection & \multirow{ 2}{*}{0.48} & \multirow{ 2}{*}{0.46} \\
        (Cohen's $\kappa$) &&\\
        \hline
        Disambiguation  & \multirow{ 2}{*}{0.08} & \multirow{ 2}{*}{0.08}\\
        (Krippendorff’s $\alpha$) &&\\
        \hline
    \end{tabular}
    \caption{Inter-annotator agreement scores. We use Cohen's Kappa for Task 1 (detection), while Krippendorff's Alpha is used for Task 2 (disambiguation).}
    \label{tab:ann_aggrement}
\end{table}

Because of the challenging nature of the annotation task, we join the annotated instances of variable mentions and link variables of each annotator.
We provide agreement scores between annotators for English and German instances separately (Table~\ref{tab:ann_aggrement}).
We calculate the Cohen's Kappa score for agreement on Variable Detection.
The scores for both English and German range between 0.46 and 0.48, which indicate that there is a moderate agreement.
For Variable Disambiguation, we use Krippendorff's Alpha.
Both languages have an agreement score of 0.08, which implies that the agreement is close to random.
One reason for such low agreement is the large number of possible variables to choose from, given that the total vocabulary size for all the documents is very large (27,365 variables that are often similar).
The annotators labeled 1,165 unique variables (around 4\% of the vocabulary) a total of 11,356 times (Table~\ref{tab:var_stats}).
In the future, we plan to analyze this disagreement with respect to the choice of variables further.

\begin{table}[]
    \centering
    \begin{tabular}{l|r}
        Type & Count  \\
        \hline
        \hline
        Total \# variables (vocab. size)  & 27,365 \\
        \hline
        \# annotated variables (tokens) & 11,356 \\
        \# uniquely used Variables (types) & 1,165\\
        \hline
    \end{tabular}
    \caption{Vocabulary size, variable types, and tokens in our SV-Ident dataset.}
    \label{tab:var_stats}
\end{table}

Looking at the document-level, variables occur with different frequency in different documents (shown in Table~\ref{tab:var_distr}).
The size of the variable vocabulary (i.e., the subset of all variables, containing only the variables from the research datasets that are linked to a publication) related to a publication ranges from 64 to 5,733.
The number of annotated variables is at least 13 and at most 1,204 for English, and for German 20 and 1,143, respectively.
The number of uniquely annotated variables is at most 153.
In the final analysis, we investigate at which ratio sentences of a document are annotated.
While the annotation ratio is at least 7\%, it is at most 86\% for relatively dense documents.\footnote{Variable-dense documents are usually short in our dataset.}

\begin{table}[]

    \centering
    \begin{tabular}{l|c|c|c|c}
         Type & \multicolumn{2}{c|}{English} & \multicolumn{2}{c}{German} \\\cline{2-5}
         & Min & Max & Min & Max \\
        \hline
        \hline
        Rel. Variables & 134 & 3,062  & 64 & 5,733\\
        \hline
        Variable tokens & 13 & 1,204  &  20 & 1,143\\
        Variable types & 4 & 153  & 5 & 54\\
        \hline
        (\%) Annt.\ Sent. & 7 & 80  & 12 & 86\\
        \hline
    \end{tabular}
    \caption{Maximum and minimum number of related variables, annotated variables, and the ratio of annotated sentences for each document in English and German.}
    \label{tab:var_distr}
\end{table}

In addition to document-level differences, variables may require contextual knowledge to be disambiguated.
Based only on the test set, annotators agree that 242 sentences had explicit, 13 implicit, and 18 both types of mentions.
At the fine-grained annotator-level, the first annotator labeled close to 37\% more implicit than explicit mentions, while the second labeled nearly thirteen times as many explicit as implicit mentions.
Given that we did not conduct calibration rounds on this specific concept, annotators may not have shared the same understanding, since this distinction was introduced only in the third round of annotations.
Future work will focus on further analyzing and validating the annotations.
We make our dataset available on GitHub as well as on HuggingFace.\footnote{\url{https://huggingface.co/datasets/vadis/sv-ident}}
In addition, we also release, as the trial dataset, the data that were originally created by \citet{zielinski-mutschke-2018-towards} (while the annotation procedure does not follow the same guideline, the data can be used as additional training data).
Notably, consecutive sentences mentioning the same variable as well as vague variable mentions were not annotated in the trial data.
We manually filter the trial data, after which, 446 English and 573 German sentences remain in the training set and 87 and 111 in the test set for each language, respectively.

\section{Experimental Setup}
\label{sec:experimental-setup}
The task of SV-Ident deals with identifying variable mentions in a text.
For simplicity, the task is formulated as a sentence-level task, but can also be solved using document-level information (in-line with the data annotation process).
The shared task is decomposed into two sub-tasks: Variable Detection and Variable Disambiguation, where the former task can be used to help filter candidate sentences for the latter.

\subsection{Tasks}
\paragraph{Task 1: Variable Detection.} The first task can be seen as a binary text classification task.
More formally, given a set of texts $T$ (in our case, sentences), for each $t$, where $t \in T$, systems should predict the binary label $l \in [0,1]$ for $t$, where a value of $1$ implies that $t$ mentions a variable.

\paragraph{Task 2: Variable Disambiguation.} The second task can be viewed as an information retrieval (IR) task, where the goal is to identify all relevant documents (i.e., variables) for a given query (i.e., input sentence).
More formally, given a set of queries $Q$ that mention variables, where $Q \subseteq T$, and the set of all documents $D$ (in our case, variables), for each $q$, where $q \in Q$, systems should predict the subset of documents $D'$ that are mentioned in $q$, where $D' \subseteq D$.

\subsection{Evaluation Metrics}
To evaluate systems, we use standard text classification and information retrieval evaluation metrics.
For the first task, systems are evaluated using the standard $F1-macro$ score averaged across languages and documents.
$F1-macro$ is defined as follows:
\begin{align}
    F1 &= \frac{1}{N}\sum^n_{n \in N}F1_n \nonumber \\
    &= \frac{1}{N}\sum^n_{n \in N} \frac{2P_nR_n}{P_n+R_n},
\end{align}
\noindent where $P$ and $R$ are the precision and recall scores, respectively.
The $F1-macro$ averages the scores for $P$ and $R$ across classes (i.e., scores are computed for each class separately and each is weighted equally).
For the second task, systems were evaluated using the (Mean) Average Precision (MAP) score with a recall cutoff value of 10 (denoted as MAP@10).
Average Precision (AP) measures the average of the precision scores at each relevant item returned (i.e., recall level) in a search result set.
MAP is the mean of the AP scores when computed across more than one query. 
MAP considers the ranking position of each relevant document.
It further assumes that a user desires to retrieve many relevant documents.
MAP is defined as follows:
\begin{align}
    \text{MAP}@K &= \frac{1}{N}\sum^n_{n \in N}AP@K_n \nonumber \\
    &= \frac{1}{N}\sum^n_{n \in N}\frac{1}{K}\sum^k_{k \in K}P@k,
\end{align}
\noindent where $P$ is the precision score, $K$ the recall level, and $N$ the number of queries.
We choose MAP over accuracy, because MAP incorporates the rank of the predicted document, which accuracy ignores.
In a realistic use-case, a user may be interested in being recommended up to $K$ relevant variables per sentence.
While we did not empirically test what value of $K$ would be most suitable for a user, we choose $K$ to equal 10, since 95\% of all sentences are labeled with up to 10 variables.
In addition to $F1-macro$ and MAP@10, we provide secondary metrics, which are not used for ranking the submitted systems, but can provide additional insights into the results.
These include precision (P), recall (R), different values of $K$ for MAP, and $R$-precision, which is the precision at recall $R$, where $R$ is the number of relevant documents for a query.

In order to account for dataset imbalance during evaluation, for each score function $f$ (i.e., evaluation metric), we compute the average score across languages and documents.
The intuition is that languages and documents are equally important, and a model should perform well on all.
The average score is computed as follows:
\begin{equation}
\label{eq:macro}
    \text{average score} = \frac{1}{L}\sum^l_{l \in L}\frac{1}{D_l}\sum^d_{d \in D_l}f(d),
\end{equation}
\noindent where $L$ is the set of languages, $D$ the set of documents, and $D_l$ the set of documents for a given language, for $l \in L$ and $D_l \subseteq D$.

\subsection{Shared Task Setup}
The shared task was hosted on CodaLab.\footnote{\url{https://codalab.lisn.upsaclay.fr/competitions/6400}}
After registering for the shared task, participants could download the test set and were asked to submit their predictions on CodaLab as a single file for each task (submissions were allowed from July 18th through August 1st, 2022).
Submissions were limited to 20 for each task.
For each submission, an automated evaluation system would upload the computed scores to the public leaderboard.

\section{Participating Systems}
\label{sec:participating-systems}
Two teams participated in our challenge on CodaLab, and one of the teams submitted a system description, which is included in the proceedings.
We summarize the report here.
The participant \citep{hoevelmeyer2022} treated both tasks, at least partly, as a problem of semantic textual similarity \citep{agirre-etal-2013-sem}.
For Task 1, sentences were first preprocessed by randomly undersampling in order to balance the data, removing stopwords, lemmatizing the data, and using only a subset of the fields from the vocabulary metadata based on preliminary experiments.
Then, test sentences and vocabulary data were converted into dense sentence representations using Sentence-T5 \citep{ni-etal-2022-sentence} for English and \texttt{Sahajtomar/German-semantic}\footnote{\url{https://huggingface.co/Sahajtomar/German-semantic}} (henceforth, GS) for German.
Similarity scores were computed for those test sentence and vocabulary item pairs.
Pairs with a score greater than a predetermined threshold were classified as sentences containing variables.
For Task 2, the same sentence representations were used, but for all test sentences.
The variables were then ranked based on their scores, with a higher score implying a greater similarity.
While other methods were also implemented, such a Logistic Regression and Multinominal Naive Bayes classifiers, the best performing systems used Sentence-T5.

\section{Evaluation}
\label{sec:evaluation}
This section first describes the baseline systems for each task and later provides the results of the shared task.

\subsection{Baselines}
\label{sec:baselines}
We train a transformer-based model for Variable Detection and implement lexical and neural zero-shot baselines for Variable Disambiguation.

The baseline system for the first task uses a transfer learning approach by fine-tuning a pre-trained language model (PLM) on the training and validation datasets.
We use a PLM that was further pre-trained on a corpus of English social science abstracts, SsciBERT \citep{shen2022sscibert}, which outperforms BERT \citep{devlin-etal-2019-bert} and SciBERT \citep{beltagy-etal-2019-scibert} models on the SV-Ident test set.
Because no multilingual or German PLM counterparts exist that have been pre-trained on scientific texts, we use the specialized monolingual SsciBERT for both English and German data.

For the second task, we implement three baseline systems in a zero-shot setting: a lexical as well as sparse and dense retrieval models.
We choose BM25 as our lexical baseline, using Elasticsearch.\footnote{\url{https://www.elastic.co/}}
For the sparse model, we use SPARTA \citep{zhao-etal-2021-sparta} and a multilingual sentence-transformer\footnote{\url{https://huggingface.co/sentence-transformers/distiluse-base-multilingual-cased-v1}} \citep{reimers-gurevych-2019-sentence,reimers-gurevych-2020-making} as the dense retriever.
Rather than training the models on the data, we use them to first encode the query and documents (i.e., variable metadata) and later rank those which are most semantically similar to a query by computing the cosine similarity between query-document pairs.
The similarity computation assumes that instances that are closer together in vector space are semantically more similar.
While participant 2 conducts an ablation study on the choice of metadata to use for matching the variables, we choose to include all metadata and leave finding the the optimal combination of metadata to future work.

\subsection{Results}
Task 1 had two participants and a single baseline system, while Task 2 had one participant and three baseline systems.
In the tables below, the systems are denoted as follows: participant 1 as Unk, participant 2 as S-T5/GS (or S-T5 for English and GS for German), the baseline for Task 1 as SSBert\textsuperscript{*}, and the baselines for Task 2 as BM25\textsuperscript{*}, Sparse\textsuperscript{*} for the SPARTA model, and Dense\textsuperscript{*} for the multilingual sentence-transformer (all baselines across text and tables are always marked with a \textsuperscript{*} asterisk).

\paragraph{Variable Detection.} 
For this task, none of the participating systems are able to beat the average score of the baseline.
Unk scores lower than chance likelihood, while, with a score of 60.17, S-T5/GS comes close to SSBert\textsuperscript{*}, which has a score of 66.10 (Table~\ref{tab:T1-Avg}).
Breaking the scores down into the average scores across documents for each language, S-T5 outperforms the baseline for English.
Thus, Task 1 can also be solved in a zero-shot setting, given that the Sentence-T5 model was not fine-tuned on the provided data.
Similar large PLMs may show further improvements.

\begin{table}[]
    \centering
    \begin{tabular}{l|l|r|r|r}
    Language &  System &    P &     R &     F1\\ %
    \hline
    \hline
      \multirow{3}{0.5em}{English} &    Unk &  65.03 &  55.24 &  38.64 \\
       &    S-T5 &  68.38 &  67.77 &  \textbf{66.96} \\
       &   SSBert\textsuperscript{*} &  \textbf{70.94} &  \textbf{70.04} &  64.28 \\
       \hline
      \multirow{3}{0.5em}{German} &    Unk &  51.66 &  52.66 &  30.88 \\
       &    GS &  59.18 &  56.04 &  53.37 \\
       &    SSBert\textsuperscript{*} &  \textbf{68.38} &  \textbf{68.53} &  \textbf{67.91} \\
       \hline
    \multirow{3}{0.5em}{Average} &    Unk &  58.35 &  53.95 &  34.76 \\
     &    S-T5/GS &  63.78 &  61.91 &  60.17 \\
     &    SSBert\textsuperscript{*} &  \textbf{69.66} &  \textbf{69.29} &  \textbf{66.10} \\
    \hline
    \end{tabular}
    \caption{Results for task 1 (detection).}
    \label{tab:T1-Avg}
\end{table}

At the document-level, systems show varying performance (see Table~\ref{tab:T1-AvgDoc} in the Appendix).
For the document with the ID 21357, participants' systems have low scores, while SSBert\textsuperscript{*} has the highest score across all documents.
Furthermore, for 7 out of the 12 documents, the baseline system has the highest score.
In addition to the number of positive and negative instances, we also report the number of variables associated with a document as well as the year of the publication.
When computing the Pearson correlation coefficient, we find a weak correlation between the $F1$ scores and the size of the search space (i.e., vocabulary size) for Unk ($r=0.132$, $p=0.68$), S-T5/GS ($r=0.266$, $p=0.26$), and SSBert\textsuperscript{*} ($r=0.152$, $p=0.64$).
With respect to the year of the document, we find a moderate correlation for Unk ($r=0.272$, $p=0.39$), S-T5/GS ($r=0.274$, $p=0.39$), and SSBert\textsuperscript{*} ($r=0.392$, $p=0.21$).
However, these correlations may not generalize due to the small number of documents.

Given the low annotator agreement with respect to the fine-grained labels, explicit and implicit, we report scores for the cases where both annotators agree on the label (see Table~\ref{tab:T1-Ann12} in the Appendix) as well as for each annotator independently (see Tables~\ref{tab:T1-Ann1} and \ref{tab:T1-Ann2} in the Appendix).
We divide the labels into \textit{explicit}, \textit{implicit}, and \textit{mixed} classes, where sentences that contain explicit and implicit variables are labeled as mixed.
In cases where both annotators agree on the label, systems perform better on explicit than on implicit or mixed mentions.
The same is true for annotator 1, except for S-T5/GS.
This implies that explicit mentions are easier to detect and disambiguate.
This is not the case for annotator 2.
A possible explanation could be the low number of implicit annotations, which may be due to a difference in understanding of the labels.
Unk outperforms all systems for the cases when both annotators agree on the label.
This is surprising given the low average performance of the system (unfortunately, no system description was provided).

\paragraph{Variable Disambiguation.}
For the second task, we report only a single submission together with the results for three baselines (Table~\ref{tab:T2-Avg}).
As described in Section~\ref{sec:evaluation}, the baselines include BM25, SPARTA (henceforth, Sparse\textsuperscript{*}), and a multilingual sentence-transformer (henceforth, Dense\textsuperscript{*}).
While we provide participants all the test sentences, we only evaluate performance on the subset of instances that contain variable mentions, as Task 1 already validates Variable Detection performance (this setup ignores false positive queries submitted by the participants).
Unless explicitly stated, the following discussion mainly focuses on the MAP@10 scores.
While the participant's system performs close to Dense\textsuperscript{*} for English, Dense\textsuperscript{*} scores twice as high for German.
Sparse\textsuperscript{*} outperforms all systems on English data.
This is likely due to the system having been trained on a large English retrieval corpus.\footnote{\url{https://github.com/microsoft/MSMARCO-Passage-Ranking}}
BM25\textsuperscript{*} and Sparse\textsuperscript{*} perform worse on German.
Lexical models, such as BM25\textsuperscript{*}, are prone to perform worse for languages that have many rare words, such as German, which allows compound nouns.
Furthermore, because Sparse\textsuperscript{*} is only specialized for English, it does not perform well for data in a different language.
Overall, Dense\textsuperscript{*} outperforms all systems by at least 0.5 points for English, except for Sparse\textsuperscript{*}, and by at least around 10 points for German.

\begin{table}[]
    \centering
    \begin{tabular}{l|l|r|r}
    Language &  System &  MAP@10 &  $R$-Prec \\
    \hline
    \hline
      \multirow{4}{*}{English} &    S-T5 &    16.27 &   14.83 \\
       &    BM25\textsuperscript{*} &   12.39 &   12.10 \\
       &    Sparse\textsuperscript{*} &   \textbf{19.02} &   \textbf{18.87} \\
       &    Dense\textsuperscript{*} &   16.96 &   15.34 \\
      \hline
      \multirow{4}{*}{German} &    GS &   10.91 &   10.35 \\
       &    BM25\textsuperscript{*} &   6.46 &    7.02 \\
       &    Sparse\textsuperscript{*} &   3.52 &    3.69 \\
       &    Dense\textsuperscript{*} &  \textbf{20.89} &   \textbf{17.96} \\
      \hline
    \multirow{4}{*}{Average} &    S-T5/GS &   13.59 &   12.59 \\
     &    BM25\textsuperscript{*} &   9.43 &   9.56 \\
     &    Sparse\textsuperscript{*} &  11.27 &  11.28 \\
     &    Dense\textsuperscript{*} &  \textbf{18.93} &  \textbf{16.65} \\
    \hline
    \end{tabular}
    \caption{Results for task 2 (disambiguation).}
    \label{tab:T2-Avg}
\end{table}

At the document-level, scores vary significantly (see Table~\ref{tab:T2-AvgDoc} in the Appendix).
Scores across different values of $K$ improve as $K$ increases.
For dense documents (i.e., documents with a high ratio of variable mention sentences), scores increase significantly when going from $k=1$ to $k=5$, such as for the IDs 21357, 57204, and 66324.
Furthermore, while some systems perform well on a document, others perform poorly.
For example, the document with ID 66324 shows the lowest performance by all systems except for BM25\textsuperscript{*}, which has a score of 22.01 and is the second-highest document score for BM25\textsuperscript{*}.
For 57561, BM25\textsuperscript{*} achieves only a score of 1.60, while all other systems score higher than 16.
S-T5/GS outperforms all baselines only once and twice when compared to only Dense\textsuperscript{*}.
Such exceptions may be caused by a larger overlap between the tokens in the document and the underlying data used to train the models.
In addition, we find a moderate correlation between MAP@10 scores and the vocabulary size (and a strong correlation for Dense\textsuperscript{*}) for S-T5/GS ($r=0.395. p=0.20$), BM25\textsuperscript{*} ($r=0.465. p=0.13$), Sparse\textsuperscript{*} ($r=0.427. p=0.17$), and Dense\textsuperscript{*} ($r=0.623. p=0.03$).
As the search space increases, performance goes down.
Finally, we find that MAP@10 is highly correlated with $R$-Precision ($r=0.941$, $p=4.99$), which implies that MAP is a good metric in the absence of the ground truth number of relevant variables.

Performance on the annotator-level is similar to that of Task 1: scores are highest when both annotators agree on the label (see Table~\ref{tab:T2-Ann12} in the Appendix).
For both annotators, scores for the explicit class are consistently higher than for either implicit or mixed classes (see Tables~\ref{tab:T2-Ann1} and \ref{tab:T2-Ann2} in the Appendix).
This means that for the task of Variable Detection, knowing whether a variable is mentioned explicitly or implicitly can mean a 10 to 20 point absolute difference in performance.
In the case when either both annotators agree on the label or when looking only at annotator 1, Sparse\textsuperscript{*} outperforms all systems.
Exploring other sparse models is a promising future direction for disambiguating implicit variable mentions.

\section{Related Work}
\label{sec:related-work}
Identifying mentions of survey variables in text was first introduced by \citet{zielinski-mutschke-2017-mining, zielinski-mutschke-2018-towards} in the OpenMinTeD project (OM).\footnote{\url{http://openminted.eu/}}
As the predecessor of our task, they created the first dataset for the problem of SV-Ident.
Table~\ref{tab:svident_vs_om} shows the statistical differences between the OM and SV-Ident datasets. Although fewer documents are annotated in SV-Ident, the number of instances in SV-Ident is almost 5 times that of OM.
To have a greater diversity of survey variables, SV-Ident corpus uses 76 datasets with more than 27k variables from different research studies, such as ALLBUS, ISSP, and Eurobarometer, whereas OM only used a single dataset.
Moreover, the SV-Ident corpus comes up with modified and additional annotation features: the unknown (UNK) token was used for ambiguous variable mentions; consecutive mentions of the same variable were included; confidence levels of the annotations and variable mention types were labeled; and variables were linked across languages.
As a result, our corpus is much larger and more diverse.

\begin{table}[]
    \centering
    \begin{tabular}{l|r|r}
         & OM & SV-Ident \\
        \hline
        \hline
        Documents & 64  &  44 \\
        \hline
        Research Datasets & 1 & 76 \\
        \hline
        Total \# variables  & \multirow{2}{*}{406} & \multirow{2}{*}{27,365} \\
        (vocabulary size) & &\\
        \hline
        \# annotated & \multirow{2}{*}{414} & \multirow{2}{*}{8,721}\\
        variables (tokens) & &\\
        \hline
        \# uniquely used & \multirow{2}{*}{243} & \multirow{2}{*}{851}\\
        variables (types) & &\\
        \hline
        Instances annotate & \multirow{2}{*}{1,217} & \multirow{2}{*}{5,972}\\
        (\# annotated sentences) & &\\
    \hline
    \end{tabular}
    \caption{Comparison between the OpenMinTeD and SV-Ident datasets.}
    \label{tab:svident_vs_om}
\end{table}

Given that identifying variables requires semantic relations, other NLP tasks deal with a fundamentally similar perspective, such as entity linking (EL), recognizing textual entailment (RTE), semantic textual similarity (STS), plagiarism detection, and detecting previously fact-checked news.
EL can be conceptualized as linking mentions to variables in a knowledge base \cite{rao2013entity}.
Since there are many similar survey variables in research datasets, disambiguating the right variable for a sentence is similar to determining the identity of an entity from a knowledge base.
The RTE task is to identify whether a sentence entails a given candidate hypothesis or not \cite{dzikovska-etal-2013-semeval}.
A question answering adaptation of RTE \cite{dagan2013recognizing} is similar to SV-Ident, as the question and each answer form a hypothesis, which then requires the system to determine whether a sentence entails a given candidate hypothesis.
STS is yet another similar task, which aims to find the similarity level between given texts \cite{agirre-etal-2013-sem}.
STS was organized as a shared International Workshop on Semantic Evaluation between 2012 and 2017, and STS models have been developed for various domains \cite{wang2020medsts,yang2020measurement,guo2020cord19sts}.
In the task of Plagiarism Detection of PAN,\footnote{\url{https://pan.webis.de/}} a system should extract all plagiarized passages from a given set of candidate documents with (external) or without (intrinsic) comparing them to potential source documents \cite{potthast2013overview}.
Lastly, Detecting Previously Fact-Checked Claims, a shared task by the CheckThat! Lab \cite{nakov2022overview}, aims to match the most similar claims --- text fragments from social media or political debate scripts --- to a corpus of verified claims.
The corpus is used to find the most similar claims, which does not require direct linking, as is done in SV-Ident Task 2, because implicit links are inferred.

\section{Why SV-Ident?}
\label{sec:why-sv-ident}

Today’s search engines are the core elements of information access for social scientists.
While search engines have seen many improvements in terms of keyword search and text understanding, they suffer from a limited capability of retrieving information from interconnected data sources, such as academic literature and research datasets.
Nonetheless, they show outstanding performance on retrieving such documents individually.
Current interlinking infrastructures typically only link research datasets to publications on the citation-level.
Such systems do not yet consider fine-grained linking of publications to individual survey variables from research datasets.
As demonstrated in the SV-Ident shared task, survey variables may be mentioned implicitly, which makes their manual or automatic identification non-trivial.
Currently, social scientists have to manually identify such variables, which is time-consuming.
In addition to these limitations, search engines do not yet support queries specific to social science topics, concepts, or relations.
Yet, keyword search, which is widely used, has many known problems (e.g., vocabulary mismatch or complex queries).
As a result, social scientists are unable to access interlinked publications and research data.
Thus, the re-use and reproducibility of research is limited.

SV-Ident, and more generally the VADIS project, plays an important role in filling the gap in the lack of infrastructure for social scientists \cite{kartal2022}.
SV-Ident aims to build automatic models for identifying survey variables in social science publications.
This directly enables a more fine-grained interlinking of publications and research datasets.
More specifically, variables can be linked on the sentence-level, which allows new features to be developed.
Within the VADIS project, we aim to develop variable-based automatic summarization, which will allow scientists to quickly get an overview of a publication with respect to the variables used.
Furthermore, we plan to incorporate variable recommendation algorithms into the GESIS Search platform to enable scientists to find relevant variables outside the scope of variables they are already familiar with.

\section{Conclusion}
\label{sec:conclusion}

This overview reports on the results of the SV-Ident 2022 shared task.
We introduce two sub-tasks relevant for SV-Ident, namely, Variable Detection and Variable Disambiguation.
We report on data, which is currently the largest of its kind, that was collected, annotated, and made publicly available for this challenge.
Baseline as well as participants' systems are described and evaluated.
We find that nearly all systems perform better on explicit variable mentions, opening up new directions of research.
Finally, we contextualize the shared task into related work and highlight its importance within a broader context.
Future work will further analyze the distinction between different variable mention types.
In addition, multi-task learning could solve both tasks jointly or in combination with adjacent tasks.
Co-reference resolution could be used to help disambiguate implicit variable mentions.
Finally, evaluating systems on more diverse metrics, such as fairness or robustness, is critical for applied research.

\paragraph{Acknowledgement}
The SV-Ident 2022 shared task is organized by the VADIS project \cite{kartal2022}. This work is supported by DFG project VADIS, grant numbers ZA 939/5-1, PO 1900/5-1, EC 477/7-1, KR 4895/3-1. 
We would like to also thank Jan Hendrik Blaß and Joudie Mekky for their contributions.

\bibliography{anthology,custom}

\begin{thebibliography}{24}
\expandafter\ifx\csname natexlab\endcsname\relax\def\natexlab#1{#1}\fi

\bibitem[{Agirre et~al.(2013)Agirre, Cer, Diab, Gonzalez-Agirre, and
  Guo}]{agirre-etal-2013-sem}
Eneko Agirre, Daniel Cer, Mona Diab, Aitor Gonzalez-Agirre, and Weiwei Guo.
  2013.
\newblock \href {https://aclanthology.org/S13-1004} {*{SEM} 2013 shared task:
  Semantic textual similarity}.
\newblock In \emph{Second Joint Conference on Lexical and Computational
  Semantics (*{SEM}), Volume 1: Proceedings of the Main Conference and the
  Shared Task: Semantic Textual Similarity}, pages 32--43, Atlanta, Georgia,
  USA. Association for Computational Linguistics.

\bibitem[{Beltagy et~al.(2019)Beltagy, Lo, and
  Cohan}]{beltagy-etal-2019-scibert}
Iz~Beltagy, Kyle Lo, and Arman Cohan. 2019.
\newblock \href {https://doi.org/10.18653/v1/D19-1371} {{S}ci{BERT}: A
  pretrained language model for scientific text}.
\newblock In \emph{Proceedings of the 2019 Conference on Empirical Methods in
  Natural Language Processing and the 9th International Joint Conference on
  Natural Language Processing (EMNLP-IJCNLP)}, pages 3615--3620, Hong Kong,
  China. Association for Computational Linguistics.

\bibitem[{Chandrasekaran et~al.(2020)Chandrasekaran, Feigenblat, Freitag,
  Ghosal, Hovy, Mayr, Shmueli-Scheuer, and
  de~Waard}]{chandrasekaran-etal-2020-overview}
Muthu~Kumar Chandrasekaran, Guy Feigenblat, Dayne Freitag, Tirthankar Ghosal,
  Eduard Hovy, Philipp Mayr, Michal Shmueli-Scheuer, and Anita de~Waard. 2020.
\newblock \href {https://doi.org/10.18653/v1/2020.sdp-1.1} {Overview of the
  first workshop on scholarly document processing ({SDP})}.
\newblock In \emph{Proceedings of the First Workshop on Scholarly Document
  Processing}, pages 1--6, Online. Association for Computational Linguistics.

\bibitem[{Dagan et~al.(2013)Dagan, Roth, Sammons, and
  Zanzotto}]{dagan2013recognizing}
Ido Dagan, Dan Roth, Mark Sammons, and Fabio~Massimo Zanzotto. 2013.
\newblock \href
  {https://doi.org/https://doi.org/10.2200/S00509ED1V01Y201305HLT023}
  {Recognizing textual entailment: Models and applications}.
\newblock \emph{Synthesis Lectures on Human Language Technologies},
  6(4):1--220.

\bibitem[{Devlin et~al.(2019)Devlin, Chang, Lee, and
  Toutanova}]{devlin-etal-2019-bert}
Jacob Devlin, Ming-Wei Chang, Kenton Lee, and Kristina Toutanova. 2019.
\newblock \href {https://doi.org/10.18653/v1/N19-1423} {{BERT}: Pre-training of
  deep bidirectional transformers for language understanding}.
\newblock In \emph{Proceedings of the 2019 Conference of the North {A}merican
  Chapter of the Association for Computational Linguistics: Human Language
  Technologies, Volume 1 (Long and Short Papers)}, pages 4171--4186,
  Minneapolis, Minnesota. Association for Computational Linguistics.

\bibitem[{Dzikovska et~al.(2013)Dzikovska, Nielsen, Brew, Leacock, Giampiccolo,
  Bentivogli, Clark, Dagan, and Dang}]{dzikovska-etal-2013-semeval}
Myroslava Dzikovska, Rodney Nielsen, Chris Brew, Claudia Leacock, Danilo
  Giampiccolo, Luisa Bentivogli, Peter Clark, Ido Dagan, and Hoa~Trang Dang.
  2013.
\newblock \href {https://aclanthology.org/S13-2045} {{S}em{E}val-2013 task 7:
  The joint student response analysis and 8th recognizing textual entailment
  challenge}.
\newblock In \emph{Second Joint Conference on Lexical and Computational
  Semantics (*{SEM}), Volume 2: Proceedings of the Seventh International
  Workshop on Semantic Evaluation ({S}em{E}val 2013)}, pages 263--274, Atlanta,
  Georgia, USA. Association for Computational Linguistics.

\bibitem[{Ejaz et~al.(2017)Ejaz, Bräuer, and Wolling}]{MaC885}
Waqas Ejaz, Marco Bräuer, and Jens Wolling. 2017.
\newblock \href {https://doi.org/10.17645/mac.v5i2.885} {Subjective evaluation
  of media content as a moderator of media effects on european identity: Mere
  exposure and the hostile media phenomenon}.
\newblock \emph{Media and Communication}, 5(2):41--52.

\bibitem[{Guo et~al.(2020)Guo, Mirzaalian, Sabir, Jaiswal, and
  Abd-Almageed}]{guo2020cord19sts}
Xiao Guo, Hengameh Mirzaalian, Ekraam Sabir, Ayush Jaiswal, and Wael
  Abd-Almageed. 2020.
\newblock \href {https://arxiv.org/abs/2007.02461} {{CORD19STS}: Covid-19
  semantic textual similarity dataset}.
\newblock \emph{arXiv preprint arXiv:2007.02461}.

\bibitem[{Hövelmeyer and Kartal(2022)}]{hoevelmeyer2022}
Alica Hövelmeyer and Yavuz~Selim Kartal. 2022.
\newblock Varanalysis@sv-ident 2022: Variable detection and disambiguation
  based on semantic similarity.
\newblock In \emph{Proceedings of the {Third} {Workshop} on {Scholarly}
  {Document} {Processing}}. Association for Computational Linguistics.

\bibitem[{Kartal et~al.(2022)Kartal, Takeshita, Tsereteli, Eckert, Kroll, Mayr,
  Ponzetto, Zapilko, and Zielinski}]{kartal2022}
Yavuz~Selim Kartal, Sotaro Takeshita, Tornike Tsereteli, Kai Eckert, Henning
  Kroll, Philipp Mayr, Simone~Paolo Ponzetto, Benjamin Zapilko, and Andrea
  Zielinski. 2022.
\newblock \href {http://arxiv.org/abs/2209.06804} {Towards {Automated} {Survey}
  {Variable} {Search} and {Summarization} in {Social} {Science}
  {Publications}}.
\newblock \emph{arXiv preprint arXiv:2209.06804}.

\bibitem[{Klie et~al.(2018)Klie, Bugert, Boullosa, Eckart~de Castilho, and
  Gurevych}]{klie-etal-2018-inception}
Jan-Christoph Klie, Michael Bugert, Beto Boullosa, Richard Eckart~de Castilho,
  and Iryna Gurevych. 2018.
\newblock \href {https://aclanthology.org/C18-2002} {The {INCE}p{TION}
  platform: Machine-assisted and knowledge-oriented interactive annotation}.
\newblock In \emph{Proceedings of the 27th International Conference on
  Computational Linguistics: System Demonstrations}, pages 5--9, Santa Fe, New
  Mexico. Association for Computational Linguistics.

\bibitem[{Nakov et~al.(2022)Nakov, Barr{\'o}n-Cede{\~n}o, da~San~Martino, Alam,
  Stru{\ss}, Mandl, M{\'\i}guez, Caselli, Kutlu, Zaghouani
  et~al.}]{nakov2022overview}
Preslav Nakov, Alberto Barr{\'o}n-Cede{\~n}o, Giovanni da~San~Martino, Firoj
  Alam, Julia~Maria Stru{\ss}, Thomas Mandl, Rub{\'e}n M{\'\i}guez, Tommaso
  Caselli, Mucahid Kutlu, Wajdi Zaghouani, et~al. 2022.
\newblock \href {https://doi.org/https://doi.org/10.1007/978-3-030-99739-7_52}
  {Overview of the clef--2022 checkthat! lab on fighting the covid-19 infodemic
  and fake news detection}.
\newblock In \emph{International Conference of the Cross-Language Evaluation
  Forum for European Languages}, pages 495--520. Springer.

\bibitem[{Ni et~al.(2022)Ni, Hernandez~Abrego, Constant, Ma, Hall, Cer, and
  Yang}]{ni-etal-2022-sentence}
Jianmo Ni, Gustavo Hernandez~Abrego, Noah Constant, Ji~Ma, Keith Hall, Daniel
  Cer, and Yinfei Yang. 2022.
\newblock \href {https://doi.org/10.18653/v1/2022.findings-acl.146}
  {Sentence-t5: Scalable sentence encoders from pre-trained text-to-text
  models}.
\newblock In \emph{Findings of the Association for Computational Linguistics:
  ACL 2022}, pages 1864--1874, Dublin, Ireland. Association for Computational
  Linguistics.

\bibitem[{Potthast et~al.(2013)Potthast, Hagen, Gollub, Tippmann, Kiesel,
  Rosso, Stamatatos, and Stein}]{potthast2013overview}
Martin Potthast, Matthias Hagen, Tim Gollub, Martin Tippmann, Johannes Kiesel,
  Paolo Rosso, Efstathios Stamatatos, and Benno Stein. 2013.
\newblock \href {https://riunet.upv.es/handle/10251/46635} {Overview of the 5th
  international competition on plagiarism detection}.
\newblock In \emph{CLEF Conference on Multilingual and Multimodal Information
  Access Evaluation}, pages 301--331. CELCT.

\bibitem[{Rao et~al.(2013)Rao, McNamee, and Dredze}]{rao2013entity}
Delip Rao, Paul McNamee, and Mark Dredze. 2013.
\newblock \href {https://doi.org/https://doi.org/10.1007/978-3-642-28569-1_5}
  {Entity linking: Finding extracted entities in a knowledge base}.
\newblock In \emph{Multi-source, multilingual information extraction and
  summarization}, pages 93--115. Springer.

\bibitem[{Reimers and Gurevych(2019)}]{reimers-gurevych-2019-sentence}
Nils Reimers and Iryna Gurevych. 2019.
\newblock \href {https://doi.org/10.18653/v1/D19-1410} {Sentence-{BERT}:
  Sentence embeddings using {S}iamese {BERT}-networks}.
\newblock In \emph{Proceedings of the 2019 Conference on Empirical Methods in
  Natural Language Processing and the 9th International Joint Conference on
  Natural Language Processing (EMNLP-IJCNLP)}, pages 3982--3992, Hong Kong,
  China. Association for Computational Linguistics.

\bibitem[{Reimers and Gurevych(2020)}]{reimers-gurevych-2020-making}
Nils Reimers and Iryna Gurevych. 2020.
\newblock \href {https://doi.org/10.18653/v1/2020.emnlp-main.365} {Making
  monolingual sentence embeddings multilingual using knowledge distillation}.
\newblock In \emph{Proceedings of the 2020 Conference on Empirical Methods in
  Natural Language Processing (EMNLP)}, pages 4512--4525, Online. Association
  for Computational Linguistics.

\bibitem[{Shen et~al.(2022)Shen, Liu, Lin, Huang, Zhang, Liu, Feng, and
  Wang}]{shen2022sscibert}
Si~Shen, Jiangfeng Liu, Litao Lin, Ying Huang, Lin Zhang, Chang Liu, Yutong
  Feng, and Dongbo Wang. 2022.
\newblock \href {https://arxiv.org/abs/2206.04510} {{SsciBERT}: A pre-trained
  language model for social science texts}.
\newblock \emph{arXiv preprint arXiv:2206.04510}.

\bibitem[{Wang et~al.(2020)Wang, Afzal, Fu, Wang, Shen, Rastegar-Mojarad, and
  Liu}]{wang2020medsts}
Yanshan Wang, Naveed Afzal, Sunyang Fu, Liwei Wang, Feichen Shen, Majid
  Rastegar-Mojarad, and Hongfang Liu. 2020.
\newblock \href {https://doi.org/https://doi.org/10.1007/s10579-018-9431-1}
  {{MedSTS}: a resource for clinical semantic textual similarity}.
\newblock \emph{Language Resources and Evaluation}, 54(1):57--72.

\bibitem[{Wilkinson et~al.(2016)Wilkinson, Dumontier, Aalbersberg, Appleton,
  Axton, Baak, Blomberg, Boiten, da~Silva~Santos, Bourne, Bouwman, Brookes,
  Clark, Crosas, Dillo, Dumon, Edmunds, Evelo, Finkers, Gonzalez-Beltran, Gray,
  Groth, Goble, Grethe, Heringa, 't~Hoen, Hooft, Kuhn, Kok, Kok, Lusher,
  Martone, Mons, Packer, Persson, Rocca-Serra, Roos, van Schaik, Sansone,
  Schultes, Sengstag, Slater, Strawn, Swertz, Thompson, van~der Lei, van
  Mulligen, Velterop, Waagmeester, Wittenburg, Wolstencroft, Zhao, and
  Mons}]{Wilkinson2016}
Mark~D. Wilkinson, Michel Dumontier, IJsbrand~Jan Aalbersberg, Gabrielle
  Appleton, Myles Axton, Arie Baak, Niklas Blomberg, Jan-Willem Boiten,
  Luiz~Bonino da~Silva~Santos, Philip~E. Bourne, Jildau Bouwman, Anthony~J.
  Brookes, Tim Clark, Merc{\`e} Crosas, Ingrid Dillo, Olivier Dumon, Scott
  Edmunds, Chris~T. Evelo, Richard Finkers, Alejandra Gonzalez-Beltran,
  Alasdair~J.G. Gray, Paul Groth, Carole Goble, Jeffrey~S. Grethe, Jaap
  Heringa, Peter~A.C 't~Hoen, Rob Hooft, Tobias Kuhn, Ruben Kok, Joost Kok,
  Scott~J. Lusher, Maryann~E. Martone, Albert Mons, Abel~L. Packer, Bengt
  Persson, Philippe Rocca-Serra, Marco Roos, Rene van Schaik, Susanna-Assunta
  Sansone, Erik Schultes, Thierry Sengstag, Ted Slater, George Strawn,
  Morris~A. Swertz, Mark Thompson, Johan van~der Lei, Erik van Mulligen, Jan
  Velterop, Andra Waagmeester, Peter Wittenburg, Katherine Wolstencroft, Jun
  Zhao, and Barend Mons. 2016.
\newblock \href {https://doi.org/10.1038/sdata.2016.18} {The fair guiding
  principles for scientific data management and stewardship}.
\newblock \emph{Scientific Data}, 3(1):160018.

\bibitem[{Yang et~al.(2020)Yang, He, Zhang, Ma, Bian, Wu
  et~al.}]{yang2020measurement}
Xi~Yang, Xing He, Hansi Zhang, Yinghan Ma, Jiang Bian, Yonghui Wu, et~al. 2020.
\newblock \href {https://doi.org/https://doi.org/10.2196/19735} {Measurement of
  semantic textual similarity in clinical texts: comparison of
  transformer-based models}.
\newblock \emph{JMIR medical informatics}, 8(11):e19735.

\bibitem[{Zhao et~al.(2021)Zhao, Lu, and Lee}]{zhao-etal-2021-sparta}
Tiancheng Zhao, Xiaopeng Lu, and Kyusong Lee. 2021.
\newblock \href {https://doi.org/10.18653/v1/2021.naacl-main.47} {{SPARTA}:
  Efficient open-domain question answering via sparse transformer matching
  retrieval}.
\newblock In \emph{Proceedings of the 2021 Conference of the North American
  Chapter of the Association for Computational Linguistics: Human Language
  Technologies}, pages 565--575, Online. Association for Computational
  Linguistics.

\bibitem[{Zielinski and Mutschke(2017)}]{zielinski-mutschke-2017-mining}
Andrea Zielinski and Peter Mutschke. 2017.
\newblock \href {https://doi.org/10.18653/v1/W17-2907} {Mining social science
  publications for survey variables}.
\newblock In \emph{Proceedings of the Second Workshop on {NLP} and
  Computational Social Science}, pages 47--52, Vancouver, Canada. Association
  for Computational Linguistics.

\bibitem[{Zielinski and Mutschke(2018)}]{zielinski-mutschke-2018-towards}
Andrea Zielinski and Peter Mutschke. 2018.
\newblock \href {https://aclanthology.org/L18-1084} {Towards a gold standard
  corpus for variable detection and linking in social science publications}.
\newblock In \emph{Proceedings of the Eleventh International Conference on
  Language Resources and Evaluation ({LREC} 2018)}, Miyazaki, Japan. European
  Language Resources Association (ELRA).

\end{thebibliography}
\bibliographystyle{acl_natbib}

\appendix
\section*{Appendix}
\label{sec:appendix}
This section contains a figure with example sentences mapped to variables and additional detailed evaluation results for both SV-Ident tasks.
More specifically, Tables~\ref{tab:T1-AvgDoc} and \ref{tab:T2-AvgDoc} provide results for each document, while Tables~\ref{tab:T1-Ann12}--\ref{tab:T1-Ann2} and Tables~\ref{tab:T2-Ann12}--\ref{tab:T2-Ann2} provide results for explicit, implicit, and mixed mention types for each annotator individually as well as for the case when both annotators agreed on the labels.

\begin{figure*}
    \centering
    \includegraphics[width=\linewidth]{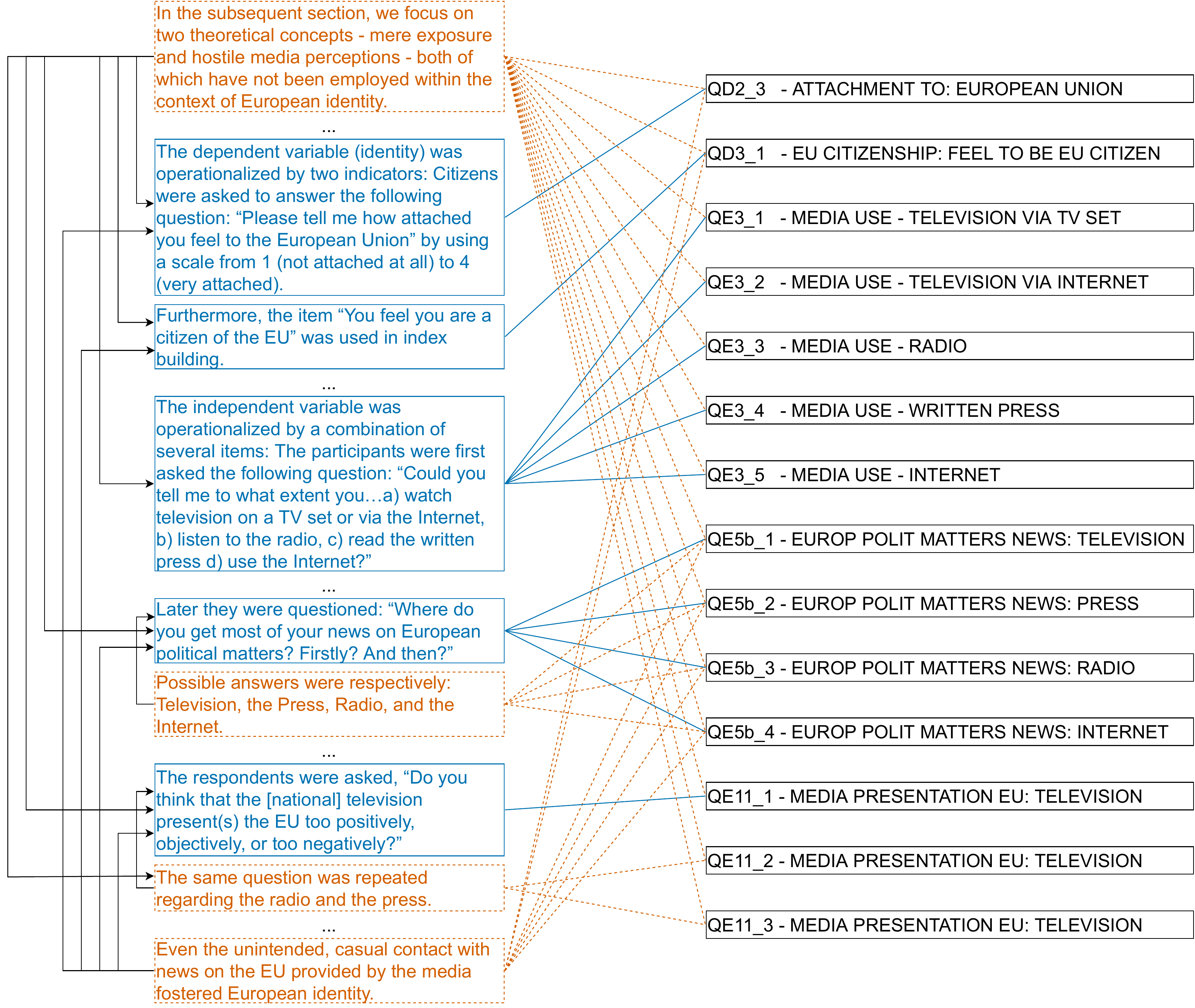}
    \caption{Example {\color[HTML]{0072B2}explicit} ({\color[HTML]{0072B2}blue}) and {\color[HTML]{D35F00}implicit} ({\color[HTML]{D35F00}red}) sentences from a social science article (source: \citet{MaC885}) mapped to survey variables. Lines with arrows show contextual dependence. Linked variables: \href{https://search.gesis.org/variables/exploredata-ZA5876_Varqd2_3}{QD2\_3}, \href{https://search.gesis.org/variables/exploredata-ZA5876_Varqd3_1}{QD3\_1}, \href{https://search.gesis.org/variables/exploredata-ZA5876_Varqe3_1}{QE3\_1}, \href{https://search.gesis.org/variables/exploredata-ZA5876_Varqe3_2}{QE3\_2}, \href{https://search.gesis.org/variables/exploredata-ZA5876_Varqe3_3}{QE3\_3}, \href{https://search.gesis.org/variables/exploredata-ZA5876_Varqe3_4}{QE3\_4}, \href{https://search.gesis.org/variables/exploredata-ZA5876_Varqe3_5}{QE3\_5}, \href{https://search.gesis.org/variables/exploredata-ZA5876_Varqe5b_1}{QE5b\_1}, \href{https://search.gesis.org/variables/exploredata-ZA5876_Varqe5b_2}{QE5b\_2}, \href{https://search.gesis.org/variables/exploredata-ZA5876_Varqe5b_3}{QE5b\_3}, \href{https://search.gesis.org/variables/exploredata-ZA5876_Varqe5b_4}{QE5b\_4}, \href{https://search.gesis.org/variables/exploredata-ZA5876_Varqe11_1}{QE11\_1}, \href{https://search.gesis.org/variables/exploredata-ZA5876_Varqe11_2}{QE11\_2}, and \href{https://search.gesis.org/variables/exploredata-ZA5876_Varqe11_3}{QE11\_3}.}
    \label{fig:large-example}
\end{figure*}

\begin{table*}[]
    \centering
    \begin{tabular}{l|r|r|r|r|c|c|c|c}
     ID &  System &    F1 &     P &     R &  \textit{\# p/n} & Vars & Lang & Year \\
    \hline
    \hline
    \multirow{3}{3em}{16547} &    Unk & 33.47 & 25.16 & 50.00 &      \multirow{3}{4em}{160/162} &    \multirow{3}{1em}{209} &        \multirow{3}{1em}{de} &         \multirow{3}{2em}{2003} \\
     &    S-T5/GS & \textbf{65.77} & \textbf{69.20} & \textbf{66.88} &       &     &         &             \\
     &    SSBert\textsuperscript{*} & 65.50 & 65.61 & 65.55 &       &     &                   &   \\
    \hline
    \multirow{3}{3em}{19944} &    Unk & 33.54 & 73.27 & 50.91 &      \multirow{3}{4em}{110/94} &    \multirow{3}{1em}{457} &        \multirow{3}{1em}{de} &            \multirow{3}{2em}{1999} \\
     &    S-T5/GS & 51.95 & \textbf{67.05} & 57.38 &       &     &         &             \\
     &    SSBert\textsuperscript{*} & \textbf{63.16} & 63.20 & \textbf{63.28} &       &     &         &             \\
    \hline
    \multirow{3}{3em}{21279} &    Unk & 22.22 & 60.00 & 52.94 &       \multirow{3}{4em}{51/12} &    \multirow{3}{1em}{477} &        \multirow{3}{1em}{de} &            \multirow{3}{2em}{1993} \\
     &    S-T5/GS & 42.61 & 42.85 & 42.40 &        &     &         &             \\
     &    SSBert\textsuperscript{*} & \textbf{67.56} & \textbf{67.08} & \textbf{68.14} &        &     &         &            \\
    \hline
    \multirow{3}{3em}{21357} &    Unk & 24.68 & 16.38 & 50.00 &       \multirow{3}{4em}{39/19} &    \multirow{3}{1em}{239} &        \multirow{3}{1em}{de} &           \multirow{3}{2em}{2002} \\
     &    S-T5/GS & 45.96 & 47.07 & 46.69 &        &     &         &             \\
     &    SSBert\textsuperscript{*} & \textbf{79.20} & \textbf{81.86} & \textbf{77.73} &        &     &         &             \\
    \hline
    \multirow{3}{3em}{21622} &    Unk & 30.43 & 59.80 & 58.16 &       \multirow{3}{4em}{49/10} &    \multirow{3}{1em}{142} &        \multirow{3}{1em}{de} &            \multirow{3}{2em}{1991} \\
     &    S-T5/GS & 63.39 & 62.30 & 65.82 &        &     &                   &   \\
     &    SSBert\textsuperscript{*} & \textbf{75.70} & \textbf{73.66} & \textbf{78.88}        &     &         &           &   \\
    \hline
    \multirow{3}{3em}{56983} &    Unk & 40.96 & 75.35 & 53.95 &       \multirow{3}{4em}{38/36} &    \multirow{3}{1em}{367} &        \multirow{3}{1em}{de} &            \multirow{3}{2em}{2018} \\
     &    S-T5/GS & 50.51 & \textbf{66.58} & 57.09 &        &     &         &             \\
     &    SSBert\textsuperscript{*} & \textbf{56.31} & 58.87 & \textbf{57.60} &        &     &         &             \\
    \hline
    \multirow{3}{3em}{49163} &    Unk & 26.12 & 64.23 & 52.06 &      \multirow{3}{4em}{97/37} &    \multirow{3}{1em}{211} &        \multirow{3}{1em}{en} &            \multirow{3}{2em}{2005} \\
     &    S-T5/GS & \textbf{63.20} & \textbf{63.06} & \textbf{65.62} &       &     &         &             \\
     &    SSBert\textsuperscript{*} & 54.27 & 62.75 & 64.38 &       &     &         &             \\
    \hline
    \multirow{3}{3em}{49734} &    Unk & 54.73 & 78.39 & 61.36 &      \multirow{3}{4em}{66/67} &    \multirow{3}{1em}{148} &        \multirow{3}{1em}{en} &            \multirow{3}{2em}{1998} \\
     &    S-T5/GS & \textbf{71.94} & 76.37 & \textbf{72.80} &       &     &         &             \\
     &    SSBert\textsuperscript{*} & 69.30 & \textbf{79.05} & 71.23 &       &     &         &             \\
    \hline
    \multirow{3}{3em}{57204} &    Unk & 31.29 & 53.07 & 50.38 &      \multirow{3}{4em}{119/77} &    \multirow{3}{1em}{134} &        \multirow{3}{1em}{en} &           \multirow{3}{2em}{2017} \\
     &    S-T5/GS & 66.45 & 67.89 & 66.08 &       &     &         &             \\
     &    SSBert\textsuperscript{*} & \textbf{66.82} & \textbf{71.17} & \textbf{70.63} &       &     &         &             \\
    \hline
    \multirow{3}{3em}{57561} &    Unk & 25.51 & 62.13 & 53.18 &      \multirow{3}{4em}{110/33} &    \multirow{3}{1em}{134} &        \multirow{3}{1em}{en} &            \multirow{3}{2em}{2017} \\
     &    S-T5/GS & \textbf{61.81} & 61.47 & 64.70 &       &     &         &             \\
     &    SSBert\textsuperscript{*} & 56.49 & \textbf{65.34} & \textbf{70.15} &       &     &         &             \\
    \hline
    \multirow{3}{3em}{61603} &    Unk & 52.92 & 66.16 & 61.92 &      \multirow{3}{4em}{71/42} &    \multirow{3}{1em}{336} &        \multirow{3}{1em}{en} &            \multirow{3}{2em}{2016} \\
     &    S-T5/GS & \textbf{77.90} & \textbf{80.75} & 76.73 &       &     &         &             \\
     &    SSBert\textsuperscript{*} & 76.84 & 78.51 & \textbf{80.23} &       &     &         &             \\
    \hline
    \multirow{3}{3em}{66324} &    Unk & 41.26 & 66.20 & 52.50 &      \multirow{3}{4em}{105/120} &    \multirow{3}{1em}{775} &        \multirow{3}{1em}{en} &          \multirow{3}{2em}{2020} \\
     &    S-T5/GS & 60.44 & 60.71 & 60.71 &       &     &         &             \\
     &    SSBert\textsuperscript{*} & \textbf{61.96} & \textbf{68.84} & \textbf{63.63} &       &     &         &             \\
    \hline
    \end{tabular}
    \caption{Fine-grained results across documents for Task 1. Sys = system, P = precision, R = recall, \# p/n = number of positive/negative sentences, Vars = total number of variables, Lang = language of the document.}
    \label{tab:T1-AvgDoc}
\end{table*}

\begin{table*}[]
    \centering
    \begin{tabular}{l|l|r|r|r|c}
    Type &     System &     F1 &      P &      R  &      \# \\
    \hline
    \hline
    \multirow{3}{4em}{A1+2exp} & Unk &  \textbf{59.45} &  \textbf{66.96} &  57.49 &  \multirow{3}{2em}{242} \\
       &  S-T5/GS &  38.16 &  53.99 &  58.73 &   \\
       &  SSBert\textsuperscript{*}  &  53.62 &  58.51 &  \textbf{70.52} &   \\
    \hline
    \multirow{3}{4em}{A1+2imp} & Unk &  \textbf{48.58} &  49.60 &  47.60 &  \multirow{3}{2em}{13} \\
       &  S-T5/GS &  25.96 &  49.92 &  47.72 &   \\
       &  SSBert\textsuperscript{*}  &  36.28 &  \textbf{50.02} &  \textbf{50.72} &   \\
    \hline
    \multirow{3}{4em}{A1+2mix} & Unk &  \textbf{48.51} &  49.45 &  47.60 &  \multirow{3}{2em}{18} \\
       &  S-T5/GS &  26.18 &  49.88 &  47.50 &   \\
       &  SSBert\textsuperscript{*}  &  36.97 &  \textbf{50.35} &  \textbf{58.28} &   \\
    \hline
    \multirow{3}{4em}{Average} & Unk &  \textbf{52.18} &  \textbf{55.34} &  50.90 &   \\
       &  S-T5/GS &  30.10 &  51.27 &  51.31 &   \\
       &  SSBert\textsuperscript{*}  &  42.29 &  52.96 &  \textbf{59.84} &   \\
    \hline
    \multirow{3}{4em}{A1+2} & Unk &  \textbf{57.66} &  \textbf{66.01} &  56.25  &  \multirow{3}{2em}{273} \\
       &    S-T5/GS  &  39.08 &  53.80 &  57.34  &   \\
        &    SSBert\textsuperscript{*}  &  54.70 &  58.88 &  68.92  &   \\
    \hline
    \end{tabular}
    \caption{Fine-grained results across types of variable mentions for Task 1. Sys = system, P = precision, R = recall, \# = number of (positive) sentences.}
    \label{tab:T1-Ann12}
\end{table*}

\begin{table*}[]
    \centering
    \begin{tabular}{l|l|r|r|r|c}
    Type &     System &     F1 &      P &      R &  \# \\
    \hline
    \hline
    \multirow{3}{3em}{A1exp} & Unk &  \textbf{57.51} &  \textbf{69.06} &  56.39 &  \multirow{3}{2em}{339} \\
      &  S-T5/GS  &  41.11 &  54.31 &  57.15 &   \\
       &  SSBert\textsuperscript{*}  &  57.05 &  60.17 &  \textbf{68.59} &   \\
    \hline
    \multirow{3}{3em}{A1imp} & Unk &  46.00 &  47.67 &  49.37 &  \multirow{3}{2em}{403} \\
      &  S-T5/GS  &  44.86 &  \textbf{55.42} &  \textbf{57.19} &   \\
       &  SSBert\textsuperscript{*}  &  \textbf{50.77} &  53.41 &  54.99 &   \\
    \hline
    \multirow{3}{3em}{A1mix} & Unk &  49.06 &  49.15 &  49.53 &  \multirow{3}{2em}{166} \\
      &  S-T5/GS  &  36.33 &  53.91 &  60.79 &   \\
       &  SSBert\textsuperscript{*}  &  \textbf{49.40} &  \textbf{56.00} &  \textbf{68.23} &   \\
    \hline
    \multirow{3}{3em}{Average} & Unk &  50.85 &  55.29 &  51.76 &   \\
      &  S-T5/GS  &  40.77 &  54.55 &  58.38 &   \\
       &  SSBert\textsuperscript{*}  &  \textbf{52.40} &  \textbf{56.53} &  \textbf{63.94} &   \\
    \hline
    \multirow{3}{3em}{A1} & Unk &  42.31 &  65.88 &  52.89 &  \multirow{3}{2em}{908} \\
      &  S-T5/GS  &  60.51 &  63.65 &  62.29 &   \\
       &  SSBert\textsuperscript{*}  &  \textbf{69.49} &  \textbf{69.58} &  \textbf{69.45} &   \\
    \hline
    \end{tabular}
    \caption{Fine-grained results across types of variable mentions for annotator 1 for Task 1. Sys = system, P = precision, R = recall, \# = number of (positive) sentences.}
    \label{tab:T1-Ann1}
\end{table*}

\begin{table*}[]
    \centering
    \begin{tabular}{l|l|r|r|r|c}
    Type &     System &     F1 &      P &      R &  \# \\
    \hline
    \hline
    \multirow{3}{3em}{A2exp} & Unk &  47.38 &  \textbf{70.97} &  54.03 &  \multirow{3}{2em}{864} \\
      &  S-T5/GS  &  58.40 &  63.71 &  63.03 &   \\
       &  SSBert\textsuperscript{*}  &  \textbf{65.45} &  65.39 &  \textbf{66.13} &   \\
    \hline
    \multirow{3}{3em}{A2imp} & Unk &  \textbf{48.82} &  49.15 &  48.60 &  \multirow{3}{2em}{74} \\ 
      &  S-T5/GS  &  27.95 &  \textbf{50.08} &  \textbf{50.66} &   \\
       &  SSBert\textsuperscript{*}  &  37.60 &  49.78 &  47.99 &   \\
    \hline
    \multirow{3}{3em}{A2mix} & Unk &  \textbf{50.51} &  50.49 &  50.59 &  \multirow{3}{2em}{74} \\
      &  S-T5/GS  &  28.93 &  50.14 &  50.82 &   \\
       &  SSBert\textsuperscript{*}  &  39.59 &  \textbf{50.66} &  \textbf{54.28} &   \\
    \hline
    \multirow{3}{3em}{Average} & Unk &  \textbf{48.91} &  \textbf{56.87} &  51.07 &   \\
      &  S-T5/GS  &  38.43 &  54.64 &  54.84 &   \\
       &  SSBert\textsuperscript{*}  &  47.55 &  55.27 &  \textbf{56.13} &   \\
    \hline
    \multirow{3}{3em}{A2}  & Unk &  44.21 &  \textbf{70.61} &  53.77 &  \multirow{3}{2em}{1012} \\
       &   S-T5/GS   &  60.37 &  63.93 &  62.62 &   \\
        &   SSBert\textsuperscript{*}   &  \textbf{65.81} &  65.82 &  \textbf{65.81} &   \\
    \hline
    \end{tabular}
    \caption{Fine-grained results across types of variable mentions for annotator 2 for Task 1. Sys = system, P = precision, R = recall, \# = number of (positive) sentences.}
    \label{tab:T1-Ann2}
\end{table*}

\begin{table*}[]
    \centering
    \begin{tabular}{l|r|r|r|r|r|r|c|c|c|c}
     ID &  System &  M@1 &  M@5 &  M@10 &  M@20 &  $R$-Prec &  \# & Vars & Lang &  Year \\
    \hline
    \hline
    \multirow{4}{3em}{16547} &    S-T5/GS &  11.55 &  14.12 &   15.20 &   16.72 &        16.55 &      \multirow{4}{2em}{160} &    \multirow{4}{1em}{209} &        \multirow{4}{1em}{de} &           \multirow{4}{2em}{2003} \\
     &    BM25\textsuperscript{*} &   2.98 &   3.30 &    3.49 &    3.57 &         3.52 &       &     &         &           \\
     &    Sparse\textsuperscript{*} &   0.03 &   0.24 &    0.51 &    0.86 &         1.34 &       &     &         &             \\
     &    Dense\textsuperscript{*} &  \textbf{20.50} &  \textbf{27.88} &   \textbf{30.41} &   \textbf{32.06} &        \textbf{31.73} &       &             &           \\
    \hline
    \multirow{4}{3em}{19944} &    S-T5/GS &   9.47 &  14.07 &   15.75 &   16.85 &        13.03 &      \multirow{4}{2em}{110} &    \multirow{4}{1em}{457} &        \multirow{4}{1em}{de} &            \multirow{4}{2em}{1999} \\
     &    BM25\textsuperscript{*} &   9.09 &  14.64 &   15.50 &   15.92 &        14.02 &       &     &         &           \\
     &    Sparse\textsuperscript{*} &   8.03 &  10.37 &   10.77 &   11.49 &        10.30 &       &     &         &           \\
     &    Dense\textsuperscript{*} &  \textbf{11.74} &  \textbf{18.41} &   \textbf{19.16} &   \textbf{19.96} &        \textbf{15.68} &       &     &         &           \\
    \hline
    \multirow{4}{3em}{21279} &    S-T5/GS &   6.65 &  11.87 &   14.47 &   16.08 &        15.94 &       \multirow{4}{2em}{51} &    \multirow{4}{1em}{477} &        \multirow{4}{1em}{de} &            \multirow{4}{2em}{1993} \\
     &    BM25\textsuperscript{*} &   4.92 &   6.96 &    7.04 &    7.24 &         7.93 &        &     &         &           \\
     &    Sparse\textsuperscript{*} &   0.00 &   1.16 &    1.63 &    2.01 &         1.91 &        &     &         &           \\
     &    Dense\textsuperscript{*} &  \textbf{14.12} &  \textbf{18.17} &   \textbf{19.88} &   \textbf{21.09} &        \textbf{18.06} &        &     &         &           \\
    \hline
    \multirow{4}{3em}{21357} &    S-T5/GS &   1.28 &   3.85 &    4.13 &    4.95 &         2.56 &       \multirow{4}{2em}{39} &    \multirow{4}{1em}{239} &        \multirow{4}{1em}{de} &            \multirow{4}{2em}{2002} \\
     &    BM25\textsuperscript{*} &   0.00 &   0.00 &    0.00 &    0.00 &         0.00 &        &     &         &           \\
     &    Sparse\textsuperscript{*} &   0.00 &   0.00 &    0.26 &    0.26 &         0.00 &        &     &         &           \\
     &    Dense\textsuperscript{*} &   \textbf{7.69} &  \textbf{15.15} &   \textbf{16.71} &   \textbf{17.12} &         \textbf{7.69} &        &     &         &           \\
    \hline
    \multirow{4}{3em}{21622} &    S-T5/GS &   3.69 &   7.05 &    9.00 &   10.15 &         7.80 &       \multirow{4}{2em}{49} &    \multirow{4}{1em}{142} &        \multirow{4}{1em}{de} &            \multirow{4}{2em}{1991} \\
     &    BM25\textsuperscript{*} &   0.34 &   3.67 &    4.01 &    4.01 &         6.07 &        &     &         &           \\
     &    Sparse\textsuperscript{*} &   1.31 &   3.49 &    3.82 &    4.21 &         4.23 &        &     &         &           \\
     &    Dense\textsuperscript{*} &   \textbf{6.63} &  \textbf{13.01} &   \textbf{14.67} &   \textbf{15.83} &        \textbf{11.18} &        &     &         &           \\
    \hline
    \multirow{4}{3em}{56983} &    S-T5/GS &   1.02 &   6.19 &    6.88 &    7.19 &         6.23 &       \multirow{4}{2em}{38} &    \multirow{4}{1em}{367} &        \multirow{4}{1em}{de} &            \multirow{4}{2em}{2018} \\
     &    BM25\textsuperscript{*} &   5.95 &   7.48 &    8.72 &   10.11 &        10.57 &        &     &         &           \\
     &    Sparse\textsuperscript{*} &   1.32 &   3.33 &    4.13 &    4.75 &         4.37 &        &     &         &           \\
     &    Dense\textsuperscript{*} &  \textbf{18.79} &  \textbf{24.06} &   \textbf{24.54} &   \textbf{25.11} &        \textbf{23.44} &        &     &         &           \\
    \hline
    \multirow{4}{3em}{49163} &    S-T5/GS &   4.33 &   8.84 &   10.28 &   11.46 &         9.16 &       \multirow{4}{2em}{97} &    \multirow{4}{1em}{211} &        \multirow{4}{1em}{en} &            \multirow{4}{2em}{2005} \\
     &    BM25\textsuperscript{*} &   1.96 &   4.32 &    5.52 &    6.05 &         3.49 &        &     &         &           \\
     &    Sparse\textsuperscript{*} &   3.89 &   7.87 &    8.92 &    9.73 &         6.61 &        &     &         &           \\
     &    Dense\textsuperscript{*} &   \textbf{8.12} &  \textbf{13.96} &   \textbf{15.46} &   \textbf{16.49} &        \textbf{11.55} &        &     &         &           \\
    \hline
    \multirow{4}{3em}{49734} &    S-T5/GS &   9.03 &  12.71 &   15.47 &   16.80 &        12.88 &       \multirow{4}{2em}{66} &    \multirow{4}{1em}{148} &        \multirow{4}{1em}{en} &            \multirow{4}{2em}{1998} \\
     &    BM25\textsuperscript{*} &  19.41 &  21.49 &   23.80 &   24.99 &        23.50 &        &     &         &           \\
     &    Sparse\textsuperscript{*} &  \textbf{23.57} &  \textbf{29.06} &   \textbf{33.57} &   \textbf{36.03} &        \textbf{30.49} &        &     &         &           \\
     &    Dense\textsuperscript{*} &  17.14 &  22.71 &   24.08 &   25.18 &        21.69 &        &     &         &           \\
    \hline
    \multirow{4}{3em}{57204} &    S-T5/GS &   7.48 &  22.12 &   31.73 &   34.41 &        31.10 &      \multirow{4}{2em}{119} &    \multirow{4}{1em}{134} &        \multirow{4}{1em}{en} &            \multirow{4}{2em}{2017} \\
     &    BM25\textsuperscript{*} &   1.21 &   5.11 &    8.51 &   13.06 &         8.80 &       &     &         &           \\
     &    Sparse\textsuperscript{*} &   \textbf{8.93} &  \textbf{28.77} &   \textbf{36.32} &   \textbf{39.48} &        \textbf{38.26} &       &     &         &           \\
     &    Dense\textsuperscript{*} &   7.09 &  16.73 &   24.90 &   30.11 &        23.82 &       &     &         &           \\
    \hline
    \multirow{4}{3em}{57561} &    S-T5/GS &   6.96 &  12.06 &   14.61 &   16.47 &        10.60 &      \multirow{4}{2em}{110} &    \multirow{4}{1em}{134} &        \multirow{4}{1em}{en} &            \multirow{4}{2em}{2017} \\
     &    BM25\textsuperscript{*} &   0.76 &   1.04 &    1.30 &    1.60 &         1.12 &       &     &         &           \\
     &    Sparse\textsuperscript{*} &   \textbf{9.33} &  13.04 &   14.82 &   16.58 &        15.70 &       &     &         &           \\
     &    Dense\textsuperscript{*} &   7.47 &  \textbf{16.80} &   \textbf{19.20} &   \textbf{20.93} &        \textbf{16.14} &       &     &         &           \\
    \hline
    \multirow{4}{3em}{61603} &    S-T5/GS &  \textbf{14.82} &  \textbf{21.32} &   \textbf{22.57} &   \textbf{23.19} &        \textbf{20.68} &       \multirow{4}{2em}{71} &    \multirow{4}{1em}{336} &        \multirow{4}{1em}{en} &            \multirow{4}{2em}{2016} \\
     &    BM25\textsuperscript{*} &  10.62 &  14.45 &   14.87 &   15.30 &        12.51 &        &     &         &           \\
     &    Sparse\textsuperscript{*} &  12.81 &  16.96 &   17.98 &   18.61 &        19.34 &        &     &         &           \\
     &    Dense\textsuperscript{*} &  12.16 &  15.47 &   16.69 &   17.19 &        16.92 &        &     &         &           \\
    \hline
    \multirow{4}{3em}{66324} &    S-T5/GS &   0.58 &   1.95 &    2.95 &    3.81 &         4.55 &      \multirow{4}{2em}{105} &    \multirow{4}{1em}{775} &        \multirow{4}{1em}{en} &            \multirow{4}{2em}{2020} \\
     &    BM25\textsuperscript{*} &   \textbf{8.17} &  \textbf{16.22} &   \textbf{20.34} &   \textbf{22.01} &        \textbf{23.18} &       &     &         &           \\
     &    Sparse\textsuperscript{*} &   0.25 &   1.92 &    2.51 &    3.15 &         2.81 &       &     &         &           \\
     &    Dense\textsuperscript{*} &   0.12 &   0.58 &    1.44 &    2.21 &         1.95 &       &     &         &           \\
    \hline
    \end{tabular}
    \caption{Fine-grained results across documents for Task 2. Sys = system, M = MAP, $R$-Prec = $R$-Precision, \# = number of (positive) sentences, Vars = total number of variables, Lang = language of the document..}
    \label{tab:T2-AvgDoc}
\end{table*}

\begin{table*}[]
    \centering
    \begin{tabular}{l|r|r|r|r|r|r|c}
       Type &  System &  M@1 &  M@5 &  M@10 &  M@20 &  $R$-Prec &  \# \\
    \hline
    \hline
    \multirow{4}{4em}{A1+2exp} &    S-T5/GS &  14.38 &  21.62 &   22.99 &   24.18 &        19.77 &      \multirow{4}{2em}{242} \\
     &    BM25\textsuperscript{*} &  13.49 &  16.00 &   16.78 &   17.17 &        16.34 &       \\
     &    Sparse\textsuperscript{*} &  11.56 &  14.57 &   15.57 &   16.23 &        14.82 &       \\
     &    Dense\textsuperscript{*} &  \textbf{24.90} &  \textbf{32.43} &   \textbf{34.31} &   \textbf{35.11} &        \textbf{30.29} &       \\
    \hline
    \multirow{4}{4em}{A1+2imp} &    S-T5/GS &   0.00 &   5.85 &    9.62 &   14.21 &        11.70 &       \multirow{4}{2em}{13} \\
     &    BM25\textsuperscript{*} &   0.00 &   4.72 &    7.79 &    9.82 &         8.46 &        \\
     &    Sparse\textsuperscript{*} &   \textbf{5.13} &  \textbf{18.07} &   \textbf{25.97} &   \textbf{27.73} &        \textbf{24.90} &        \\
     &    Dense\textsuperscript{*} &   1.54 &   6.22 &   11.85 &   15.15 &         9.42 &        \\
    \hline
    \multirow{4}{4em}{A1+2mix} &    S-T5/GS &   6.07 &  11.10 &   13.74 &   15.55 &        16.13 &       \multirow{4}{2em}{18} \\
     &    BM25\textsuperscript{*} &   0.00 &   1.73 &    1.94 &    2.72 &         4.48 &        \\
     &    Sparse\textsuperscript{*} &   4.03 &  11.63 &   15.14 &   18.51 &        18.31 &        \\
     &    Dense\textsuperscript{*} &   \textbf{8.05} &  \textbf{14.60} &   \textbf{15.84} &   \textbf{18.66} &        \textbf{20.09} &        \\
    \hline
    \hline
       \multirow{4}{4em}{Average} &    S-T5/GS &   6.81 &  12.85 &   15.45 &   17.98 &        15.87 &        \\
        &    BM25\textsuperscript{*} &   4.50 &   7.48 &    8.84 &    9.91 &         9.76 &         \\
        &    Sparse\textsuperscript{*} &   6.90 &  14.75 &   18.89 &   20.82 &        19.34 &        \\
        &    Dense\textsuperscript{*} &  \textbf{11.50} &  \textbf{17.75} &   \textbf{20.67} &   \textbf{22.97} &        \textbf{19.93} &        \\
    \hline
    \hline
       \multirow{4}{4em}{A1+2} &    S-T5/GS &  12.92 &  19.91 &   21.52 &   22.95 &        19.03 &      \multirow{4}{2em}{273} \\
        &    BM25\textsuperscript{*} &  11.68 &  14.25 &   15.12 &   15.63 &        14.97 &       \\
        &    Sparse\textsuperscript{*} &  10.61 &  14.53 &   16.12 &   17.05 &        15.66 &       \\
        &    Dense\textsuperscript{*} &  \textbf{22.27} &  \textbf{29.57} &   \textbf{31.60} &   \textbf{32.70} &        \textbf{28.32} &       \\
    \hline
    \end{tabular}
    \caption{Fine-grained results across types of variable mentions for Task 2. Sys = system, M = MAP, $R$-Prec = $R$-Precision, \# = number of (positive) sentences.}
    \label{tab:T2-Ann12}
\end{table*}

\begin{table*}[]
    \centering
    \begin{tabular}{l|r|r|r|r|r|r|c}
     Type &  System &  M@1 &  M@5 &  M@10 &  M@20 &  $R$-Prec &  \# \\
    \hline
    \hline
    \multirow{4}{3em}{A1exp} &    S-T5/GS &  14.14 &  20.47 &   21.87 &   22.77 &        16.97 &      \multirow{4}{2em}{271} \\
     &    BM25\textsuperscript{*} &  13.91 &  15.98 &   16.76 &   17.22 &        15.27 &       \\
     &    Sparse\textsuperscript{*} &  11.77 &  14.84 &   15.61 &   16.25 &        13.92 &       \\
     &    Dense\textsuperscript{*} &  \textbf{25.28} &  \textbf{31.48} &   \textbf{32.60} &   \textbf{33.30} &        \textbf{28.05} &       \\
    \hline
    \multirow{4}{3em}{A1imp} &    S-T5/GS &   3.44 &   7.65 &   10.83 &   12.33 &         9.23 &      \multirow{4}{2em}{370} \\
     &    BM25\textsuperscript{*} &   2.23 &   4.73 &    5.98 &    7.27 &         4.93 &       \\
     &    Sparse\textsuperscript{*} &   \textbf{4.79} &  \textbf{10.63} &   \textbf{13.06} &   \textbf{14.15} &        \textbf{12.55} &       \\
     &    Dense\textsuperscript{*} &   3.89 &   8.97 &   11.82 &   13.78 &        10.20 &       \\
    \hline
    \multirow{4}{3em}{A1mix} &    S-T5/GS &   2.41 &   5.20 &    7.24 &    8.59 &         7.69 &      \multirow{4}{2em}{153} \\
     &    BM25\textsuperscript{*} &   3.63 &   6.10 &    6.98 &    7.87 &         6.80 &       \\
     &    Sparse\textsuperscript{*} &   3.16 &   6.64 &    7.45 &    8.68 &         8.32 &       \\
     &    Dense\textsuperscript{*} &   \textbf{4.85} &  \textbf{10.12} &   \textbf{11.88} &   \textbf{13.60} &        \textbf{11.89} &       \\
    \hline
    \hline
     \multirow{4}{2em}{Average} &    S-T5/GS &   6.66 &  11.10 &   13.32 &   14.57 &        11.30 &         \\
      &    BM25\textsuperscript{*} &   6.59 &   8.94 &    9.91 &   10.79 &         9.00 &          \\
      &    Sparse\textsuperscript{*} &   6.57 &  10.70 &   12.04 &   13.03 &        11.60 &         \\
      &    Dense\textsuperscript{*} &  \textbf{11.34} &  \textbf{16.86} &   \textbf{18.77} &   \textbf{20.23} &        \textbf{16.71} &         \\
    \hline
    \hline
       \multirow{4}{2em}{A1} &    S-T5/GS &   6.89 &  11.55 &   13.91 &   15.18 &        11.58 &      \multirow{4}{2em}{794} \\
        &    BM25\textsuperscript{*} &   6.49 &   8.83 &    9.85 &   10.78 &         8.82 &       \\
        &    Sparse\textsuperscript{*} &   6.86 &  11.30 &   12.85 &   13.81 &        12.20 &       \\
        &    Dense\textsuperscript{*} &  \textbf{11.37} &  \textbf{16.87} &   \textbf{18.93} &   \textbf{20.41} &        \textbf{16.62} &       \\
    \hline
    \end{tabular}
    \caption{Fine-grained results across types of variable mentions for annotator 1 for Task 2. Sys = system, M = MAP, $R$-Prec = $R$-Precision, \# = number of (positive) sentences.}
    \label{tab:T2-Ann1}
\end{table*}

\begin{table*}[]
    \centering
    \begin{tabular}{l|r|r|r|r|r|r|c}
     Type &  System &  M@1 &  M@5 &  M@10 &  M@20 &  $R$-Prec &  \# \\
    \hline
    \hline
    \multirow{4}{2em}{A2exp} &    S-T5/GS &  10.38 &  16.74 &   19.48 &   20.89 &        16.57 &      \multirow{4}{2em}{637} \\
     &    BM25\textsuperscript{*} &   7.73 &  11.14 &   12.74 &   13.81 &        11.12 &       \\
     &    Sparse\textsuperscript{*} &   7.29 &  12.55 &   14.17 &   15.02 &        12.24 &       \\
     &    Dense\textsuperscript{*} &  \textbf{15.90} &  \textbf{23.47} &   \textbf{26.25} &   \textbf{27.66} &        \textbf{21.65} &       \\
    \hline
    \multirow{4}{2em}{A2imp} &    S-T5/GS &   0.00 &   3.22 &    5.41 &    6.38 &         4.30 &       \multirow{4}{2em}{48} \\
     &    BM25\textsuperscript{*} &   1.04 &   3.39 &    4.50 &    5.89 &         4.09 &        \\
     &    Sparse\textsuperscript{*} &   3.24 &   7.61 &    9.20 &   10.47 &         6.91 &        \\
     &    Dense\textsuperscript{*} &   \textbf{7.52} &  \textbf{12.87} &   \textbf{14.43} &   \textbf{14.98} &        \textbf{10.14} &        \\
    \hline
    \multirow{4}{3em}{A2mix} &    S-T5/GS &   3.99 &   9.58 &   13.96 &   15.11 &        13.90 &       \multirow{4}{2em}{74} \\
     &    BM25\textsuperscript{*} &   2.44 &   4.65 &    5.26 &    6.62 &         6.42 &        \\
     &    Sparse\textsuperscript{*} &   5.24 &  13.03 &   15.93 &   17.54 &        16.35 &        \\
     &    Dense\textsuperscript{*} &   \textbf{7.94} &  \textbf{13.35} &   \textbf{15.91} &   \textbf{18.28} &        \textbf{18.00} &        \\
    \hline
    \hline
     \multirow{4}{2em}{Average} &    S-T5/GS &   4.79 &   9.85 &   12.95 &   14.13 &        11.59 &         \\
      &    BM25\textsuperscript{*} &   3.74 &   6.39 &    7.50 &    8.77 &         7.21 &          \\
      &    Sparse\textsuperscript{*} &   5.26 &  11.06 &   13.10 &   14.34 &        11.83 &         \\
      &    Dense\textsuperscript{*} &  \textbf{10.45} &  \textbf{16.56} &   \textbf{18.86} &   \textbf{20.31} &        \textbf{16.59} &         \\
    \hline
    \hline
       \multirow{4}{2em}{A2} &    S-T5/GS &   9.14 &  15.23 &   18.09 &   19.44 &        15.55 &      \multirow{4}{2em}{753} \\
        &    BM25\textsuperscript{*} &   6.82 &  10.06 &   11.54 &   12.65 &        10.25 &       \\
        &    Sparse\textsuperscript{*} &   6.85 &  12.28 &   14.01 &   14.96 &        12.27 &       \\
        &    Dense\textsuperscript{*} &  \textbf{14.65} &  \textbf{21.88} &   \textbf{24.56} &   \textbf{26.01} &        \textbf{20.58} &       \\
    \hline
    \end{tabular}
    \caption{Fine-grained results across types of variable mentions for annotator 2 for Task 2. Sys = system, M = MAP, $R$-Prec = $R$-Precision, \# = number of (positive) sentences.}
    \label{tab:T2-Ann2}
\end{table*}

\end{document}